\documentclass[conference]{IEEEtran}
\IEEEoverridecommandlockouts
\usepackage{cite}
\usepackage{amsmath,amssymb,amsfonts}
\usepackage{algorithmic}
\usepackage{graphicx}
\usepackage{textcomp}
\usepackage{xcolor}
\usepackage{array}
\usepackage{hyperref}
\usepackage{enumerate}
\usepackage{multirow}
\usepackage{subfigure}
\usepackage{diagbox}

\usepackage{floatrow}
\newfloatcommand{capbtabbox}{table}[][\FBwidth]

\def\BibTeX{{\rm B\kern-.05em{\sc i\kern-.025em b}\kern-.08em
    T\kern-.1667em\lower.7ex\hbox{E}\kern-.125emX}}
\begin{document}
	
\bibliographystyle{IEEEtran}
	
\title{A Preliminary Study on Data Augmentation of Deep Learning for Image Classification}

\author{\IEEEauthorblockN{Benlin Hu, Cheng Lei, Dong Wang, Shu Zhang, Zhenyu Chen$^*$}
	\IEEEauthorblockA{\textit{State Key Lab of Novel Software Technology, Nanjing University, China} \\
		 $^*$zychen@nju.edu.cn}}
\maketitle 

\begin{abstract}
Deep learning models have a large number of free parameters that need to be calculated by effective training of the models on a great deal of training data to improve their generalization performance. However, data obtaining and labeling is expensive in practice. Data augmentation is one of the methods to alleviate this problem. In this paper, we conduct a preliminary study on how three variables (augmentation method, augmentation rate and size of basic dataset per label) can affect the accuracy of deep learning for image classification. The study provides some guidelines: (1) it is better to use transformations that alter the geometry of the images rather than those just lighting and color. (2) 2-3 times augmentation rate is good enough for training. (3) the smaller amount of data, the more obvious contributions could have.
\end{abstract}


\section{Introduction} \label{sec:introduction}

Deep learning is powerful, but they usually need to be trained on massive amounts of data to perform well, which can be considered as a major limitation\cite{zhou2006training}\cite{DBLP:journals/corr/abs-1801-01078}. Deep learning\cite{lecun2015deep} models trained on small dataset show the low performance of versatility and generalization from the validation and test set. Hence, these models suffer from the problem caused by over-fitting. 

A quantity of methods have been proposed to reduce the over-fitting problem\cite{wang2017regularization}. Data augmentation, which increases both the amount and diversity of data by ``augmenting" them, is an effective strategy that we can reduce over-fitting on models and improve the diversity of the dataset and generalization performance\cite{NIPS2012_4824}. In the application of image classification, there have already been some general augmentations, like flipping the image horizontally or vertically, translating the image by a few pixels and so on.

In order to have an in-depth investigation and propose a guideline about how to use the augmentation methods properly, we perform a preliminary experiment to summarize guidelines and testify the explanation of the phenomenon. In the experiment, 10 state of the art augmentation methods are studied on two popular datasets, which representing gray images and color images.

The datasets used in the experiment are the MNIST\cite{lecun1998mnist} and CIFAR-10\cite{krizhevsky2009learning}. MNIST consists of 60000 handwritten digit images in the training set and 10000 in the test set, which are in gray-scale with 10 classes with image dimensions of $28\times28\times1$. CIFAR-10 contains 50000 training and 10000 testing $32\times32\times3$ color images with 10 classes as well.

The main contributions of this paper are summarized as follows. A preliminary experiment has been conducted to find out how the accuracy of a deep learning model can be affected by the three variables: \textbf{augmentation method}, \textbf{augmentation rate} and \textbf{size of basic dataset per label}. According to the experimental results, some guidelines based on the three variables have been summarized to use augmentation methods properly. The further experimental results show that the simple augmentation methods, such as altering the geometry of the image, have better effectiveness than the complicated ones.

\section{Data Augmentation}

\begin{figure*}[htbp]
    \centering
    \caption{Examples of augmentation methods}
    \label{fig:augment_sample}
    \subfigure[Translate]{
      \includegraphics[width=.16\textwidth]{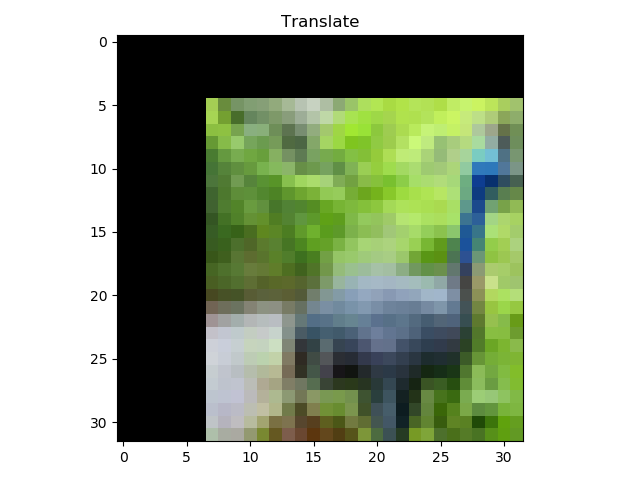}
    }
    \quad
    \subfigure[Scale]{
      \includegraphics[width=.16\textwidth]{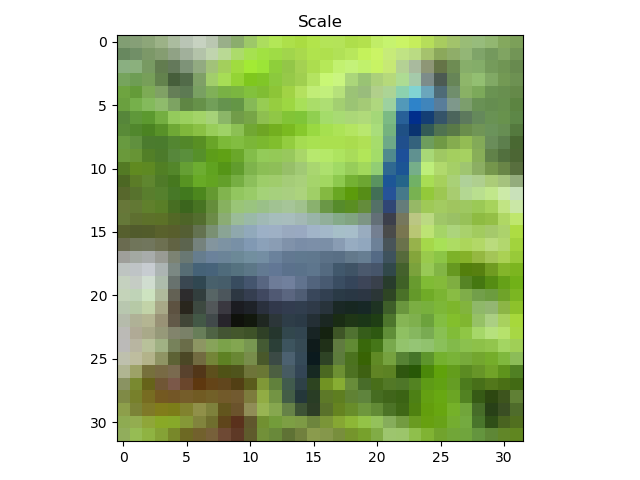}
    }
    \quad
    \subfigure[Rotate]{
      \includegraphics[width=.16\textwidth]{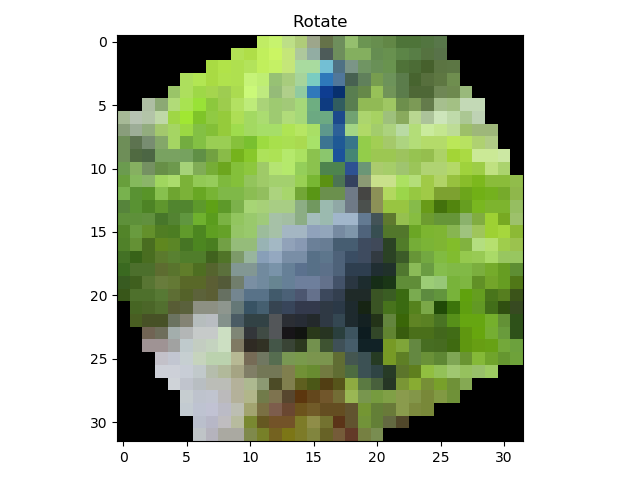}
    }
    \quad
    \subfigure[Shear]{
      \includegraphics[width=.16\textwidth]{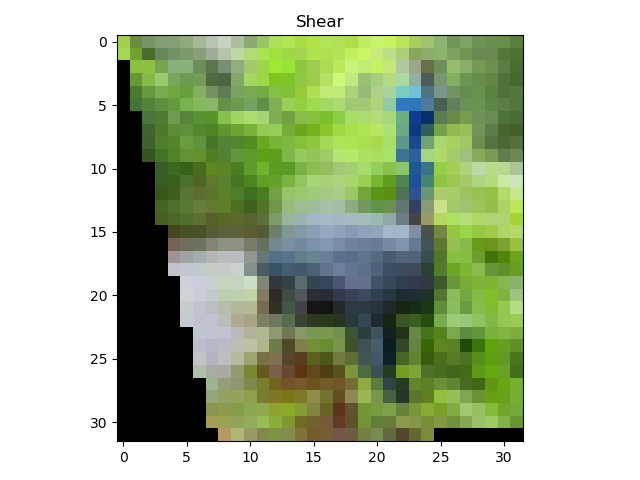}
    }
    \quad
    \subfigure[Invert]{
      \includegraphics[width=.16\textwidth]{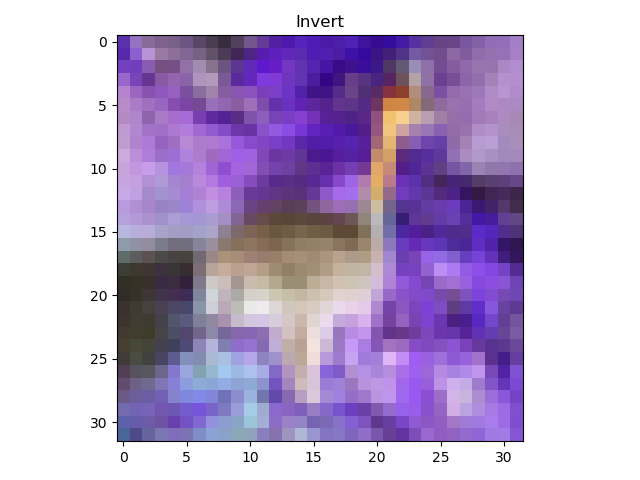}
    }
    \quad
    \subfigure[Solarize]{
      \includegraphics[width=.16\textwidth]{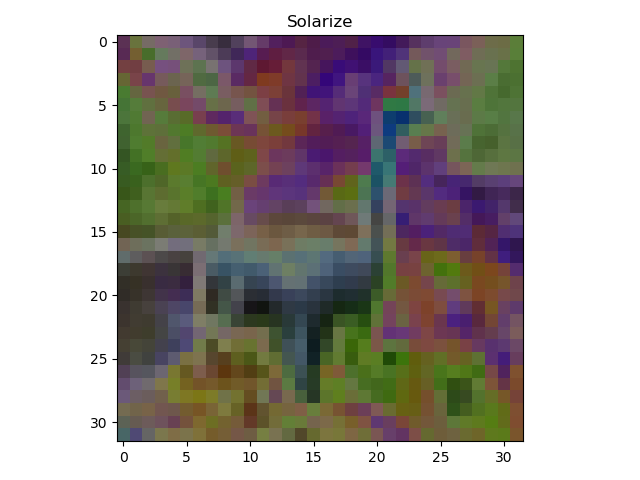}
    }
    \quad
    \subfigure[Equalize]{
      \includegraphics[width=.16\textwidth]{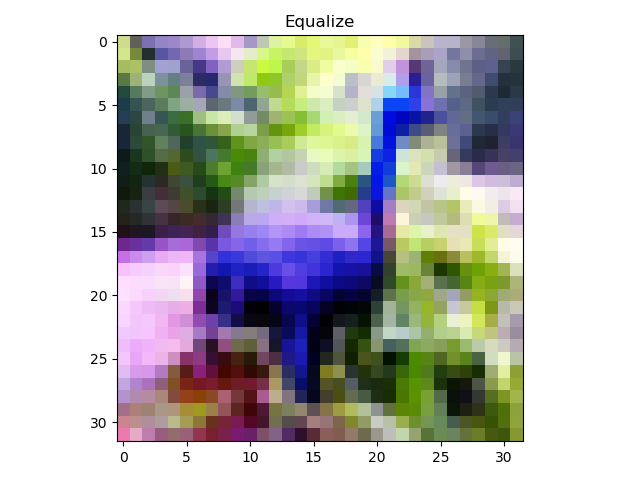}
    }
    \quad
    \subfigure[Color balance]{
      \includegraphics[width=.16\textwidth]{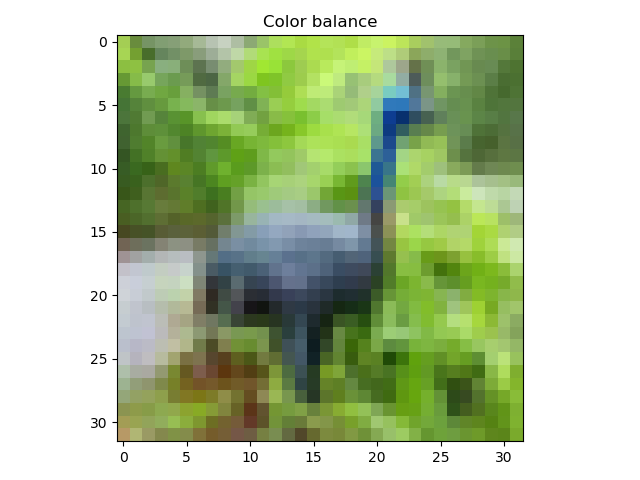}
    }
    \quad
    \subfigure[Auto contrast]{
      \includegraphics[width=.16\textwidth]{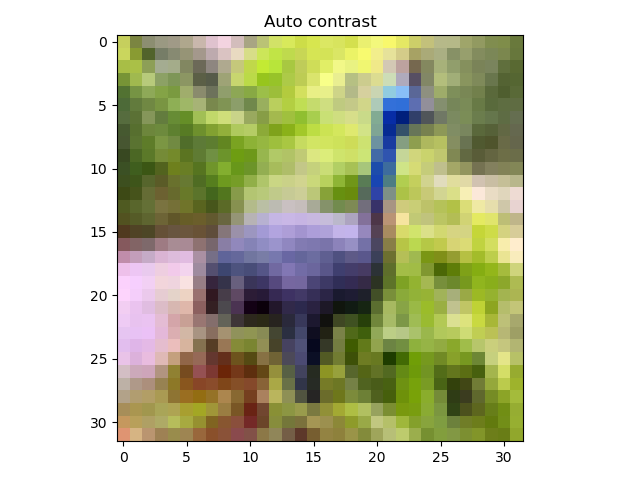}
    }
    \quad
    \subfigure[Cutout]{
      \includegraphics[width=.16\textwidth]{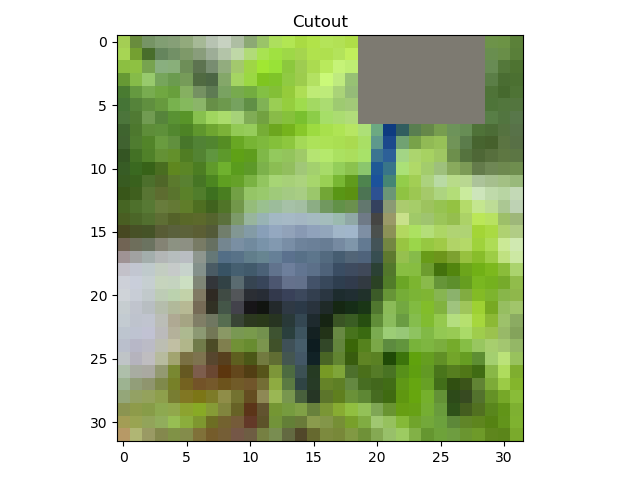}
    }
\end{figure*}

Based on the existing work, 10 state of the art augmentation methods, shown in Fig. \ref{fig:augment_sample}, for image classification are adopted in the experiment. As for the selection of experimental models, ResNet-20 and LeNet-5 are chosen for CIFAR-10 and MNIST to conduct our experiment, respectively. Table \ref{tab:augmentation_methods} lists the methods and their own descriptions and ranges we used to augment data. 

The reason why different augmentation methods can improve the model performance is because they mimic the image with different \textit{features} when taking photos. Furthermore, these augmentation methods have different effects in improving accuracy, because there are differences existing in the quantity and quality of these \textit{features} in the datasets. Therefore, it is valuable to study how these \textit{features} can influence on the model performance.

\begin{table}[!t]
    \centering
    \newcommand{\linebreakcell}{\multicolumn{2}{p{4cm}|}}
    \begin{scriptsize}
    \caption{Augmentation methods in experiment}
    \renewcommand{\arraystretch}{1.3} 
    \begin{tabular}{|l|ll|l|}
        \hline
        \textbf{Name} & \linebreakcell{\textbf{Description}} & \textbf{Range} \\
        \hline
        \textit{Translate} & \linebreakcell{Translate the image in the horizontal and vertical direction with rate $magnitude$} & $[-0.3, 0,3]$ \\
        \hline
        \textit{Scale} & \linebreakcell{Zoom in or Zoom out the image with rate $magnitude$ and select the center of the scaled image} & $[-0.5, 0.5]$ \\
        \hline
        \textit{Rotate} & \linebreakcell{Rotate the image $magnitude$ degrees} & $[-30^{\circ}, 30^{\circ}]$ \\
        \hline
        \textit{Shear} & \linebreakcell{Shear the image along the horizontal or vertical axis with rate $magnitude$} & $[-0.3, 0.3]$ \\
        \hline
        \textit{Invert} & \linebreakcell{Invert the pixels of the image} &  \\
        \hline
        \textit{Solarize} & \linebreakcell{Invert all pixels above a threshold value of $magnitude$} & $[0, 256]$ \\
        \hline
        \textit{Equalize} & \linebreakcell{Equalize the image histogram} &  \\
        \hline
        \textit{Color balance} & \linebreakcell{Adjust the color balance of the image. A $magnitude$ = 0 gives a black and white image, while $magnitude$ = 1 gives the original image} & $[0.1, 1.9]$ \\
        \hline
        \textit{Auto contrast} & \linebreakcell{Maximize the contrast of the image} &  \\
        \hline
        \textit{Cutout} & \linebreakcell{Set a random square patch of side-length $magnitude$ pixels to gray} & $[0, 20]$ \\
        \hline
    \end{tabular}
    \label{tab:augmentation_methods}
    \end{scriptsize}
\end{table}

The original dataset with $c$ classes is denoted as $M$. The basic dataset used to train model is defined as $N \subseteq M$, and the size of dataset per label in the basic dataset $N$ is $cnum$. $f$ and $rate$ are the augmentation method and its rate we chose, respectively. The dataset, which is augmented from $N$ by the augmentation method $f$, is named augmented dataset $N'$. Obviously, the size of basic dataset $|N|$ is $c*cnum$ and the size of augmented dataset $|N'|$ is $c*cnum*rate$. The model , which are trained by the basic dataset $N$ and augmented dataset $N'$, are denoted as $m_{basic}$ and $m_{augmented}$, respectively. The accuracy of $m_{basic}$ is defined as $acc_{basic}$, and the counterpart of $m_{augmented}$ is $acc_{augmented}$. The augmented test dataset is denoted as $T'$, which is augmented from test dataset $T$ by the augmentation method $f$. $acc_{aug\_test}$ is the accuracy of the $m_{basic}$ evaluated by augmented test dataset $T'$. And we use \textit{feature rate}, defined as $(acc_{basic} - acc_{aug\_test})$, to represent the \textit{features} that the model $m_{basic}$ has not yet learned from the basis dataset $N$.

\section{Experiment}

\subsection{Experiment Design}
Our experiment is designed as follows:
\begin{enumerate}
    \item The training set $N$ is randomly chosen from $M$ in accordance with $cnum$. The basic dataset $N$ is used to train the model $m_{basic}$. 
    \item On the basis of $N$, each data is augmented $rate$ times by the augmentation methods $f$.
    \item A new model $m_{augmented}$ is trained on the augmented dataset $N'$. 
    \item The model $m_{basic}$ is evaluated with accuracy $acc_{aug\_test}$ by using augmented test dataset $T'$. 
    \item The improvement of accuracy and the \textit{feature rate} of the model are used as evaluation criteria.
\end{enumerate}

In order to obtain convincing results without loss of generality, each experiment is repeated 10 times, and the average values are taken as the final experimental results.

\subsection{Experimental Results}

\begin{figure*}[htbp]
    \centering
    \caption{The experimental results of LeNet-5 on MNIST}
    \label{fig:mnist_ex_result}
    \subfigure[The size of basic dataset: 50]{
      \includegraphics[width=.3\textwidth]{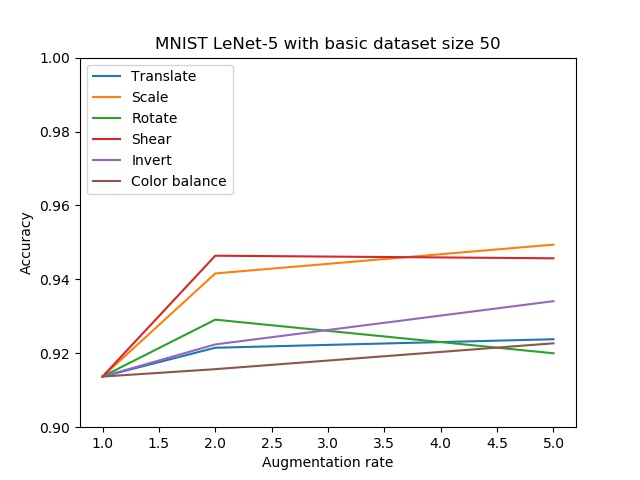}
    }
    \quad
    \subfigure[The size of basic dataset: 100]{
      \includegraphics[width=.3\textwidth]{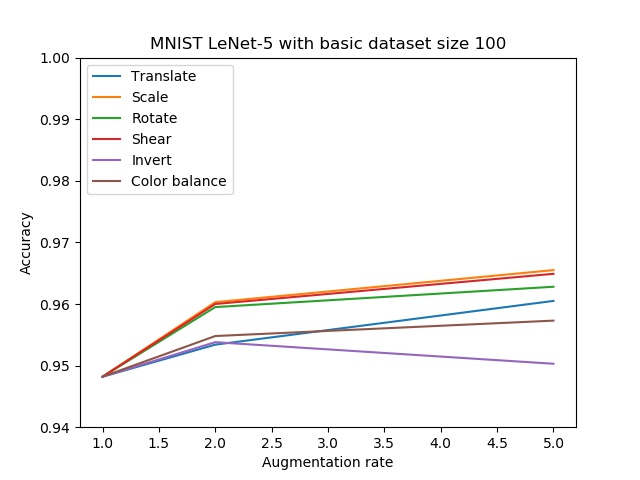}
    }
    \quad
    \subfigure[The size of basic dataset: 200]{
      \includegraphics[width=.3\textwidth]{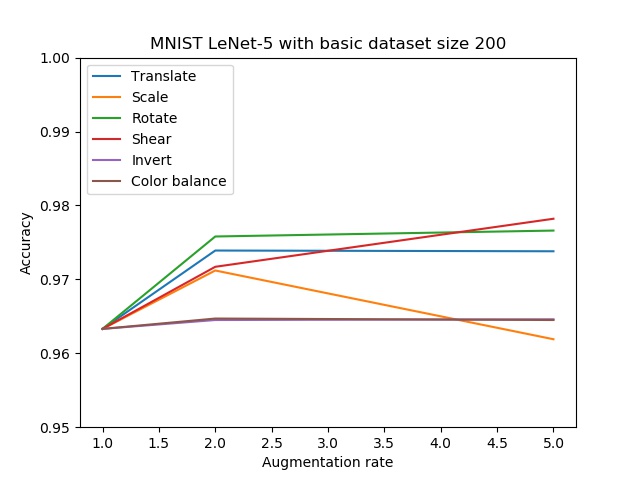}
    }
    \quad
    \subfigure[The size of basic dataset: 300]{
      \includegraphics[width=.3\textwidth]{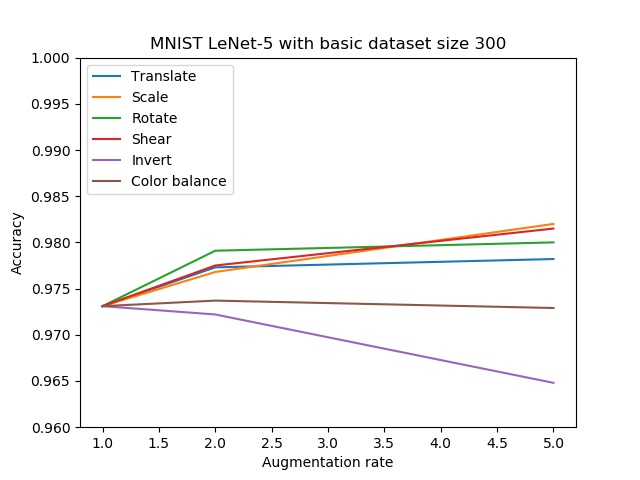}
    }
    \quad
    \subfigure[The size of basic dataset: 400]{
      \includegraphics[width=.3\textwidth]{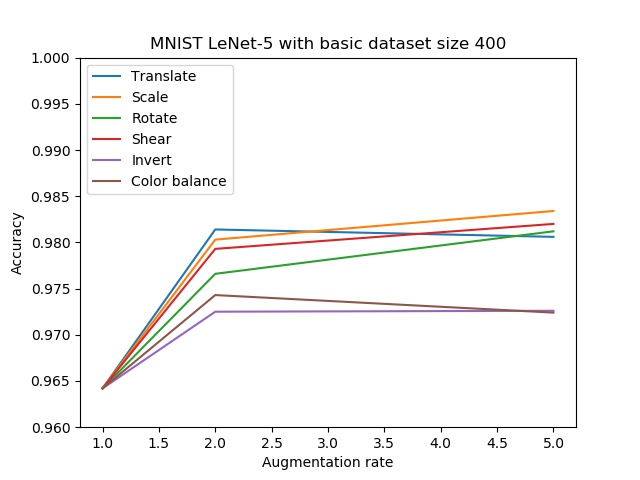}
    }
    \quad
    \subfigure[The size of basic dataset: 500]{
      \includegraphics[width=.3\textwidth]{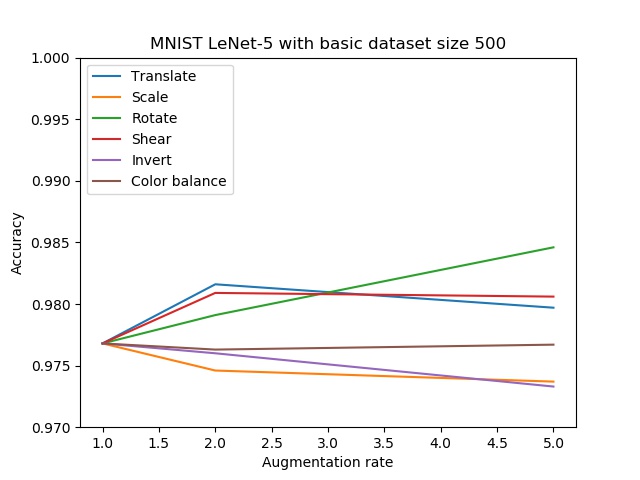}
    }
\end{figure*}

\begin{figure*}[htbp]
    \centering
    \caption{The experimental results of ResNet-20 on CIFAR-10}
    \label{fig:cifar10_ex_result}
    \subfigure[The size of basic dataset: 2000]{
      \includegraphics[width=.3\textwidth]{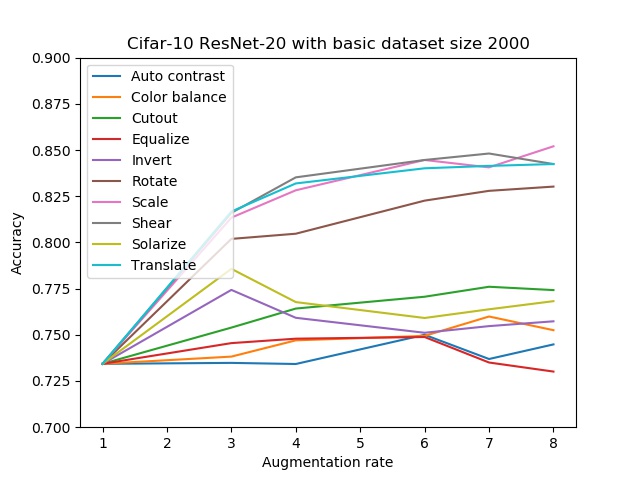}
    }
    \quad
    \subfigure[The size of basic dataset: 3000]{
      \includegraphics[width=.3\textwidth]{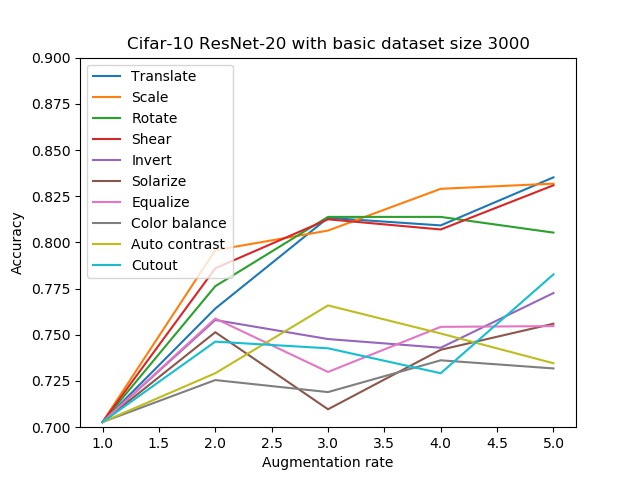}
    }
    \quad
    \subfigure[The size of basic dataset: 4000]{
      \includegraphics[width=.3\textwidth]{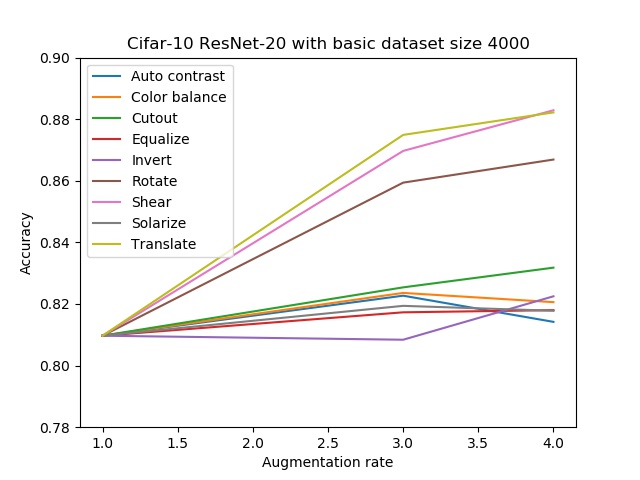}
    }
\end{figure*}

The experimental results are shown in Fig. \ref{fig:mnist_ex_result} and  \ref{fig:cifar10_ex_result}. For MNIST, the three methods of \textit{Translate}, \textit{Scale} and \textit{Rotate} are good and relatively stable, while the \textit{Shear} method is good but unstable. For CIFAR-10, almost all augmentation methods can improve accuracy. Among them, the \textit{Rotate}, \textit{Scale}, \textit{Shear} and \textit{Translate} have outstanding performance, which is 6\%-11\% higher than the control group. The augmentation methods are roughly sorted in descending order of improvement as follows: (1) \textit{Translate}, \textit{Scale}, \textit{Shear}. (2) \textit{Rotate}. (3) \textit{Cutout}, \textit{Solarize}. (4) \textit{Invert}, \textit{Equalize}, \textit{Auto Contrast}, \textit{Color balance}.

The accuracy improvement is used to evaluate the performance of different augmentation methods. The average values and medians of the improvement are listed in Table \ref{tab:mnist_accuracy_improvement} and Table \ref{tab:cifar10_accuracy_improvement}. When it comes to a small amount of data, the augmentation methods can still work well. The smaller the amount of data, the more obvious the contribution it has, which indicates that data augmentation has a large space and potential in the field where learnable data is scarce, such as rare disease diagnosis and large earthquake prediction.

\begin{table}[htbp]
    \begin{scriptsize}
    \begin{tabular}{|c||c|c|c|c|c|}
    \hline
    method           & translate & scale & rotate & shear & invert \\
    \hline
    $fea.$ rate avg. & 0.7695 & 0.1598 & 0.0791 & 0.4741 & 0.5045 \\
    \hline
    impro. avg.      & 0.0067 & 0.0110 & 0.0055 & 0.0120 & 0.0046 \\
    \hline
    impro. median    & 0.0061 & 0.0078 & 0.0086 & 0.0079 & 0.0034 \\
    \hline
    \end{tabular}
    \caption{The feature rate and the accuracy improvement on MNIST}
    \label{tab:mnist_accuracy_improvement}
    \end{scriptsize}
\end{table}

\begin{table}[htbp]
    \begin{scriptsize}
    \begin{tabular}{|c||c|c|c|c|c|}
    \hline
    method           & translate & scale & rotate & shear & invert \\
    \hline
    $fea.$ rate avg. & 0.2731 & 0.2640 & 0.2153 & 0.2175 & 0.3722 \\
    \hline
    impro. avg.      & 0.1294 & 0.1339 & 0.1038 & 0.1348 & 0.0821 \\
    \hline
    impro. median    & 0.1084 & 0.1149 & 0.0945 & 0.1098 & 0.0500 \\
    \hline
    \hline
    method           & solarize & equalize & col. bal. & auto con. & cutout \\
    \hline
    $fea.$ rate avg. & 0.1599 & 0.1396 & 0 & 0.0133 & 0.063 \\
    \hline
    impro. avg.      & 0.0617 & 0.0467 & 0.0224 & 0.0299 & 0.0469 \\
    \hline
    impro. median    & 0.0438 & 0.0516 & 0.0194 & 0.0343 & 0.0417 \\
    \hline
    \end{tabular}
    \caption{The feature rate and the accuracy improvement on CIFAR-10}
    \label{tab:cifar10_accuracy_improvement}
    \end{scriptsize}
\end{table}

Most augmentation methods can improve the accuracy by increasing the augmentation rate. However, the time, space and other costs increase rapidly, but the improvement of accuracy increases slowly. Small-scale experiments with higher augmentation rate also show that this is not cost-effective. In practice, unless there are extremely stringent requirements for accuracy, it is generally not necessary to take 5 times or more for a little improvement. 

The results show that the simpler the augmentation method, the more obvious the improving effectiveness. This confirms the assumption when using a more complex model that it is better to use transformations that alter the geometry of the images rather than just lighting and coloring\cite{DBLP:journals/corr/abs-1708-06020}.

\begin{figure}[htbp]
    \centering
    \caption{The correlations between feature rate and the accuracy improvement}
    \label{fig:feature_rate_acc_improve_result}
    \subfigure[Lenet-5 on MNIST]{
      \includegraphics[width=.4\textwidth]{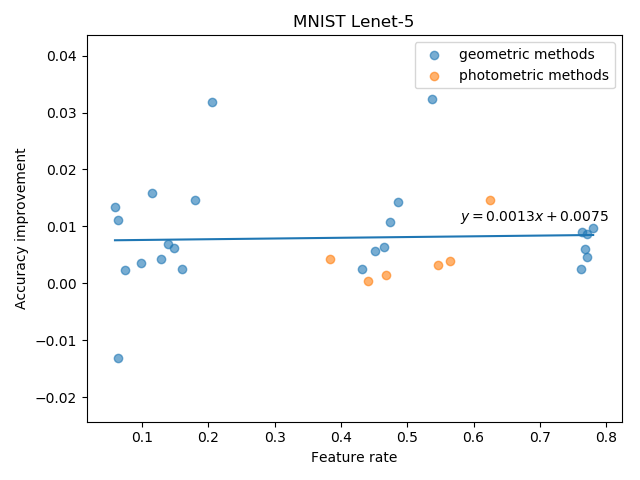}
    }
    \quad
    \subfigure[ResNet-20 on CIFAR-10]{
      \includegraphics[width=.4\textwidth]{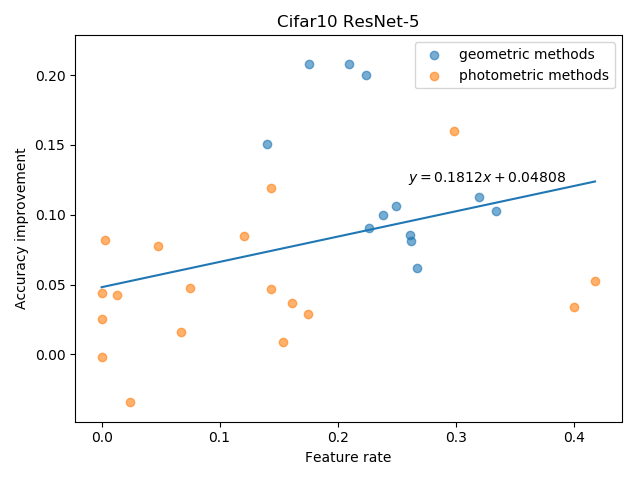}
    }
\end{figure}

Figure \ref{fig:feature_rate_acc_improve_result} shows the results of positive correlation between the \textit{feature rate} and the accuracy improvement on both datasets, especially on CIFAR-10.  High \textit{feature rate} means that the model has not learned enough \textit{features} of images when it was trained on the basic dataset $N$. And being trained on augmented dataset $N'$ provides the model more \textit{features}, thus the accuracy of the model improves. In the results of CIFAR-10, geometric methods generally better than photometric methods. This may inspire us that \textit{features} could be one of the reasons why it is better to use transformations that alter the geometry of the images rather than just lighting and coloring.

\section{Related Work} \label{sec:related_work}

Image processing methods are implemented by PIL which accept an image as input and output a processed image\cite{DBLP:journals/corr/abs-1805-09501}. The methods include ShearX/Y, TranslateX/Y, Rotate, AutoContrast, Invert, Equalize, Solarize, Posterize, Contrast, Color, Brightness and Sharpness, as shown in Figure \ref{fig:augment_sample}.

Various geometrical and photometric schemes are evaluated on a coarse-grained dataset using a relatively simple CNN\cite{DBLP:journals/corr/abs-1708-06020}. The experimental results indicate that, under these circumstances, \textit{cropping} in \textit{geometric augmentation} significantly increases CNN task performance.

Affine transformation, a 2D geometric transform method, is based on reflecting the image, scaling and translating the image, and rotating the image by different angles. The affine augmentation method is very common and widely used for correcting geometric distortion introduced by perspective\cite{stearns1995method}.

Cutout is originally considered as a targeted method for removing visual features with high activations in later layers of a convolutional neural network (CNN). However, the results in \cite{DBLP:journals/corr/abs-1708-04552}  \cite{DBLP:journals/corr/abs-1708-04896} show that randomly selecting a rectangle region in an image and erasing its pixels with random values can be used to improve the overall performance of CNNs.

Histogram equalization is introduced as a data augmentation method\cite{7784231}. Histogram equalization, solarization and adjusting image color balance are common methods used in digital image processing. These methods can simulate the problems encountered when taking photos improperly, like harsh lighting combined with auto-white-balance will produce images that over or under exposed.  

There are many existing studies discussed data augmentation methods. However, many of them focus on new data augmentation methods. The empirical study in this paper provides some guidelines to use data augmentation methods properly according to the application scenario.

\section{Conclusion and Future Work} \label{sec:conclusion_and_future_work}

Data augmentation methods generate new data by performing image processing methods such as rotation and translation on the training set. Therefore, these methods expand the training set in the amount and generalization degree. Our experimental results show that these augmentation methods work well even on small dataset. Applying higher augmentation rate is not cost-effective, because the marginal benefit is gradually reduced while the time, space and other costs increase linearly. Our study recommends that the best augmentation rate is 2-3 times. The simple augmentation methods such as translation, rotation, scaling and shearing can achieve good results, and in fact, much better than more complicated methods. 

The results show that the performance of professional image processing methods such as \textit{Equalize} and \textit{Auto Contrast} is poor. A possible reason is that \textit{features} pointed to by the simple method are more common in most images.  However, models trained with training set which augmented by simple methods can get more learning data and stronger generalization ability, because an augmentation method means adding more \textit{features} to the data, and then the model can eliminate the interference caused by the \textit{features} and perform better. Visualization\cite{zhang2019neuralvis} may be used to understand the operating principle of augmentation methods.

In this paper, due to the consideration of time budget, the study is preliminary. Larger datasets and more complicated models will be studied in the future. There are some opportunities to improve effectiveness via combining different single augmentation methods. The mutation methods may be another strategy to \cite{shen2018munn} improve data augmentation. 
The data augmentation methods can also be considered in test set to reduce the labeling cost\cite{shi2019deepgini}. 

\section{Acknowledgement}
The work is partly supported by the National Natural Science Foundation of China (61832009, 61802171).

\bibliographystyle{IEEEtran}
\bibliography{cite}

\end{document}